\documentclass[journal,twoside,web]{ieeecolor}

\usepackage{jsen}
\usepackage{cite}
\usepackage{caption}
\usepackage{amsmath,amssymb,amsfonts}
\usepackage{algorithmic}
\usepackage{graphicx}
\usepackage{textcomp}
\usepackage{wrapfig}
\usepackage{booktabs}
\usepackage{graphicx}
\usepackage{float}
\usepackage{subfig}
\usepackage{flushend}
\usepackage{xcolor}

\usepackage[normalem]{ulem} 

\def\BibTeX{{\rm B\kern-.05em{\sc i\kern-.025em b}\kern-.08em
		T\kern-.1667em\lower.7ex\hbox{E}\kern-.125emX}}
\definecolor{abstractbg}{rgb}{0.89804,0.94510,0.83137}
\begin{document}
	\title{Reverse Attitude Statistics Based Star Map Identification Method}
	\author{Shunmei Dong, Qinglong Wang, Haiqing Wang, and Qianqian Wang* 
		\thanks{Shunmei Dong and Qianqian Wang are with the School of Optics and Photonics, Beijing Institute of Technology, Beijing 100081, China (e-mail: shunmei.dong@bit.edu.cn;qqwang@bit.edu.cn).(Corresponding author: Qianqian Wang.)}
		\thanks{Qinglong Wang and Haiqing Wang are with the Beijing Institute of Control and Electronic Technology, Beijing 100032, China (e-mail:kself@163.com;wanghq@egaid.com.cn).}}
	
	\IEEEtitleabstractindextext{%
		\fcolorbox{abstractbg}{abstractbg}{%
			\begin{minipage}{\textwidth}%
				\begin{wrapfigure}[14]{r}{3.3in}%
					\includegraphics[width=3.2in]{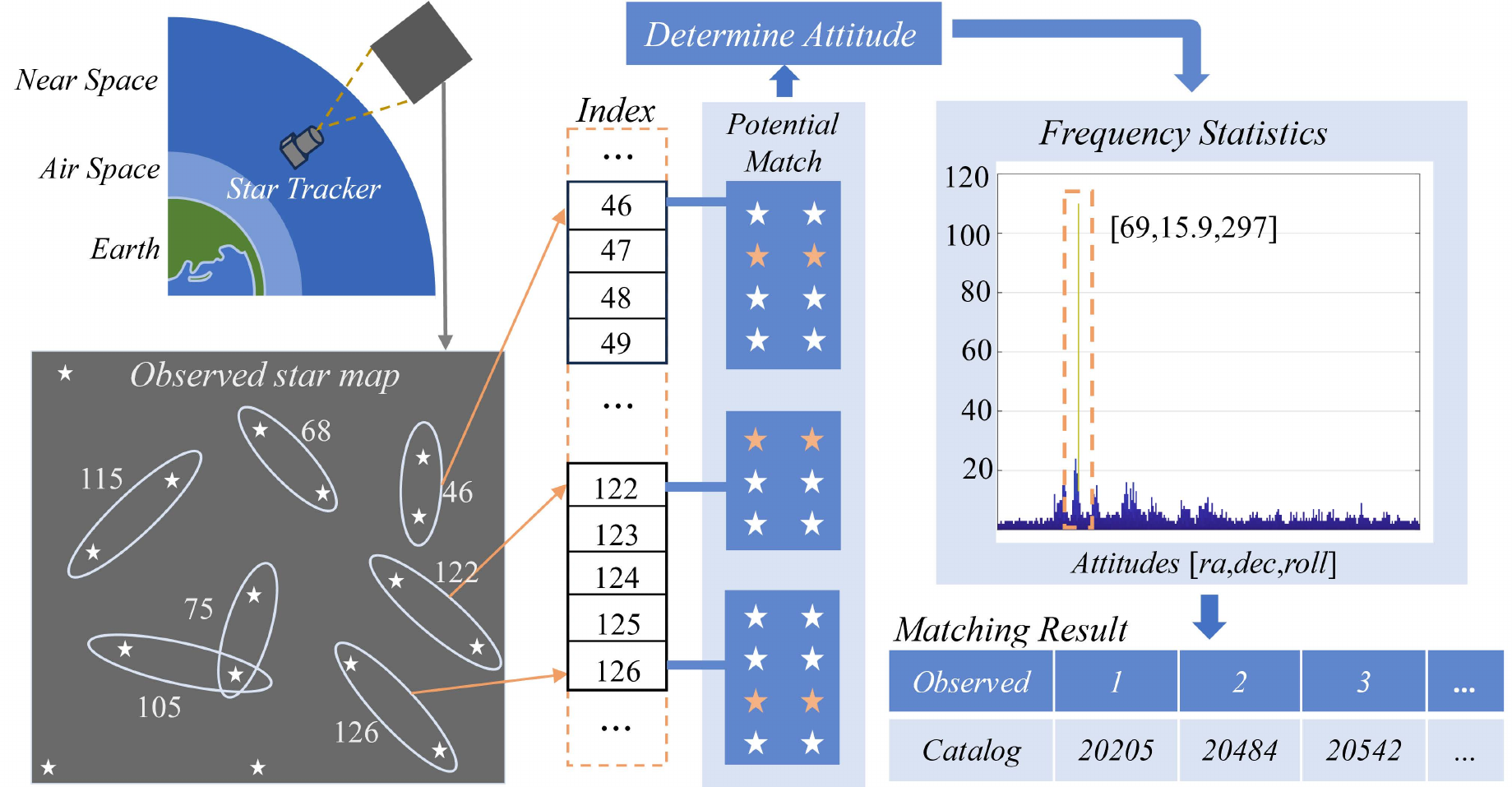}%
				\end{wrapfigure}%
				\begin{abstract}
					The star tracker is generally affected by the atmospheric background light and the aerodynamic environment when working in near space, which results in missing stars or false stars. Moreover, high-speed maneuvering may cause star trailing, which reduces the accuracy of the star position.
					To address the challenges for star map identification, a reverse attitude statistics based method is proposed to handle position noise, false stars, and missing stars. Conversely to existing methods which match before solving for attitude, this method introduces attitude solving into the matching process, and obtains the final match and the correct attitude simultaneously by frequency statistics.
					Firstly, based on stable angular distance features, the initial matching is obtained by utilizing spatial hash indexing. Then, the dual-vector attitude determination is introduced to calculate potential attitude. Finally, the star pairs are accurately matched by applying a frequency statistics filtering method.
					In addition, Bayesian optimization is employed to find optimal parameters under the impact of noises, which is able to enhance the algorithm performance further.
					In this work, the proposed method is validated in simulation, field test and on-orbit experiment. Compared with the state-of-the-art, the identification rate is improved by more than 14.3\%, and the solving time is reduced by over 28.5\%.
				\end{abstract}
				
	\end{minipage}}}
	
	\maketitle
	\section{Introduction}
	\label{sec:introduction}
	\IEEEPARstart{T}{he} star tracker has become the preferred attitude-sensitive device for spacecraft because of its excellent precision, significant autonomy, and non-accumulation of attitude errors over time. Nowadays, with the advancement of technology, lightweight star trackers have the potential to be applied in near space.
	Its further applications face challenges of the atmospheric background light during daytime and the aerodynamic environment caused by maneuvers of the spacecraft.
	Under the combined effect of these two factors, the light of the imaging background is bright, and some of the stars are buried or appear as false stars caused by the atmospheric flow field.
	These bring difficulties to the identification rate and robustness of current methods.
	
	Existing star map identification methods are classified into subgraph isomorphism, pattern recognition, and artificial intelligence.
	
	In subgraph-isomorphism-based algorithms, a catalog star map is typically constructed by all the stars in the database. To obtain features of the star map,  
	consider each star as a vertex and the angular distances between them as edges.
	These algorithms regard the observed image as a subgraph of the catalog star map.
	The identification process is matching the most similar one based on those features \cite{fan2023multi,dai2024robust,ramos2023star,wei2024star,niu2022fast}.
	The most representative approach is the triangle algorithm proposed by Liebe \textit{et al.} \cite{liebe1993pattern}. Three stars and their angular distances are used to generate a triangle feature, which is simple to achieve. However, the search process is seriously time-consuming since the catalog triangle star database is extensive.
	To enhance the efficiency and noise immunity of the algorithm, Mortari \textit{et al.} proposed a pyramid algorithm \cite{mortari2004pyramid}.
	Based on the triangle algorithm, the fourth star is introduced, and the pyramid subgraph is formed by replacement, which reduces the redundancy and improves the identification accuracy. Although the algorithm is robust to false stars, its performance degrades as their number increases.
	To improve the speed, Zhang \textit{et al.} proposed an improved triangle algorithm for fast retrieval of catalog star maps by constructing a partition table \cite{zhang2006yi}.
	Zhao \textit{et al.} and Wang \textit{et al.} utilized K-L transform and hash map to reduce the dimensionality of geometric features, respectively, which effectively reduces the amount of data, but the robustness is inadequate \cite{zhao2016star,wang2017star}.
	Kolomenkin \textit{et al.} proposed a method based on geometric voting to enhance the robustness. This method uses the principle of most similar angular distance to vote for the correspondence between the catalog and observed stars. The highest votes are considered as the matching output \cite{kolomenkin2008geometric}.
	Moreover, Yuan \textit{et al.} proposed a global voting-based method, which applies principal component analysis (PCA) to reduce the dimensionality of the triangle features before performing global voting \cite{yuan2022star}. However, the voting-based method is sensitive to false stars and has heavy computational burdens, making it difficult to fulfill the real-time requirement of the spacecraft.
	
	For the pattern-recognition-based approaches, the positional distribution of stars is utilized to generate a unique pattern, which is compared with a pre-calculated database to find the most similar one \cite{shen2024hybrid,fu2024maximum,liao2024redundant}.
	The grid algorithm proposed by Padgett \textit{et al.} is a typical method, which maps the star to a grid plane \cite{padgett1997grid}, thereby matching the patterns. The grid algorithm has an excellent identification rate, but it is affected by neighboring stars and requires numerous observed stars. Therefore, Na \textit{et al.} \cite{na2009modified} and Yoon \textit{et al.} \cite{yoon2013autonomous} improved it to enhance the identification efficiency, but it still requires massive real stars.
	In addition, Silani \textit{et al.} proposed the Polestar algorithm based on radial features \cite{silani2006star}. However, the distribution information of stars is insufficiently utilized, causing the algorithm to be vulnerable to position noise. To strengthen the robustness, an algorithm based on radial and cyclic features is designed by Zhang \textit{et al.} \cite{zhang2008full}. But the selection of the starting position may be disturbed by false stars.
	
	Artificial-intelligence-based methods employ star patterns as learning features and apply neural networks, deep learning \textit{etc.}\cite{rijlaarsdam2020efficient,xu2019rpnet,zhu2023autonomous}. 
	Hong \textit{et al.} proposed an autonomous identification algorithm adopting fuzzy neural logic networks \cite{hong2000neural}.
	The all-sky star map identification problem is solved by Paladugu \textit{et al.} \cite{paladugu2006intelligent} employing an improved genetic algorithm, where a primary star is selected, and the angular distance between stars is calculated and encoded as features. Yang \textit{et al.} adopted a one-dimensional convolutional neural network (CNN) with combined star \cite{yang2022robust}, and hybrid features are extracted for alignment before identification. Wang \textit{et al.} proposed to construct star features via log-polar coordinate transformation, then apply a CNN for classification \cite{wang2021efficient}. Artificial intelligence-based methods are complex in structure, which is inappropriate for engineering applications. Furthermore, the amount of data processing is enormous, and the hardware requirements are high.
	
	Reviewing the state-of-the-art, the pattern-recognition-based and artificial-intelligence-based approaches are unreliable. Their sensitivity to false and missing stars makes it hard to apply them in complex optical environments in near space. Although subgraph-isomorphism-based algorithms have remarkable robustness, they also have a drawback of slow identification speed.
	To address the above problems, a novel star map identification method is proposed in this work. The original contributions are:
	1) A reverse-attitude-statistics-based framework is proposed to identify star maps, which utilizes attitude-solving results to assist matching. By adopting angular distance hash indexing and attitude frequency statistics, the identification rate and robustness are enhanced while the time consumption is reduced simultaneously.
	2) Bayesian optimization is employed to find optimal parameters, which is able to improve the algorithm performance under the impact of noises;
	3) The simulation, field test, and on-orbit experiment are designed to validate the proposed method, and the robustness and real-time of the algorithm are analyzed.
	
	The rest of this paper is organized as follows. In Section \uppercase\expandafter{\romannumeral2}, the framework of the proposed method is described. Section \uppercase\expandafter{\romannumeral3} presents a comparison of the proposed approach with state-of-the-art in simulation, field test and on-orbit experiment, respectively. Finally, conclusions are drawn in Section \uppercase\expandafter{\romannumeral4}.
	
	\section{Method Description}
	\subsection{Overall Architecture}
\begin{figure}[tb]
		\centering
		\includegraphics[width=0.9\columnwidth]{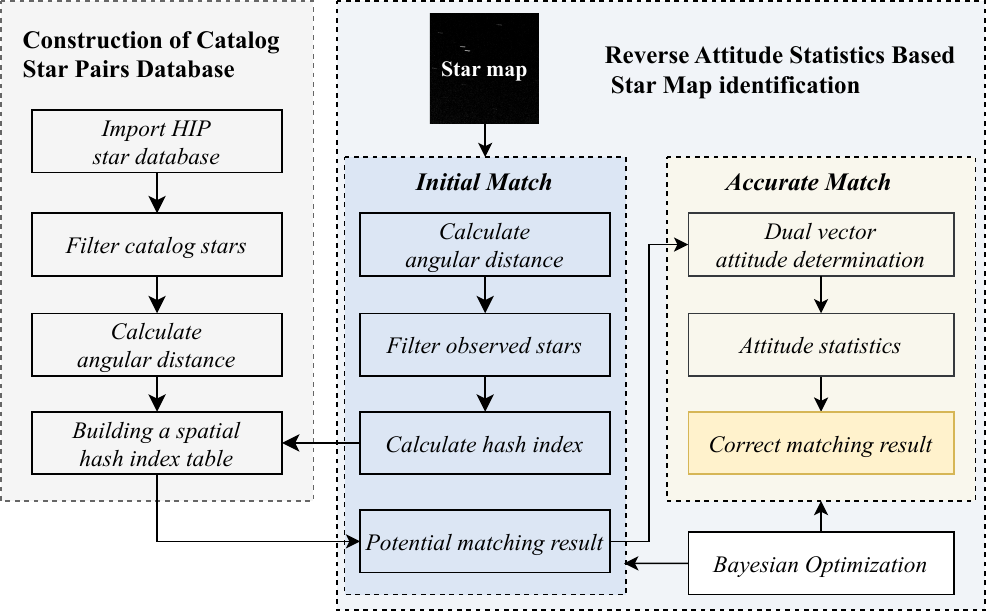}
		\caption{The framework of reverse attitude statistics based star map identification method.}
		\label{fig1}
	\end{figure}

	The overall architecture of our proposed method is shown in Fig.(\ref{fig1}).
	In this work we proposed a novel and reliable star map identification method to determine the HIP (Hipparcos Catalogues) numbers of key stars in the camera field of view (FOV), which in turn provides the necessary information for the star sensor to calculate the current attitude. It is a identification method based on star pair angular distance and consists of two parts: 1) constructing the catalog star pair database. The HIP star database is imported and filtered by magnitude, then angular distances are calculated, and recorded by utilizing a hash map; 2) performing the star map identification based on reverse attitude statistics, which is divided into two steps. The initial match includes calculating the angular distance of observed stars, filtering the star pairs in FOV, and calculating index values to find potential matches in the hash map. In the process of accurate match, the attitudes of the star tracker are calculated by a dual-vector-based algorithm and analyze the frequency to obtain the correct matching result. Moreover, Bayesian optimization is adopted in these two steps to enhance the efficiency and improve the robustness.
	
	\subsection{Construction of Catalog Star Pairs Database}
	\subsubsection{Angular Distance Features}
	The task of star map identification is to find a set of catalog stars whose distribution pattern is sufficiently similar to that of the observed star map.
	The angular distance feature is a stable geometric feature with translational and rotational invariance. Moreover, the value of angular distance is insensitive to lighting variations, magnitude errors, image noise, \textit{etc.}, which makes it reliable for star map identification.
	The direction vectors of catalog stars $i$ and $j$ in the celestial spherical inertial coordinate system are denoted as $cs_{i}$ and $cs_{j}$, respectively. And $os_{1}$ and $os_{2}$ represent the direction vectors of observed stars $1$ and $2$ in the sensor coordinate system, respectively. The angular distances of them are calculated as \cite{zhang2011xing}:
	
	\begin{subequations}\label{dr,dm}
		\begin{align}
			{d_r}^{ij} = \arccos (\frac{{c{s_i} \cdot c{s_j}}}{{|c{s_i}| \cdot |c{s_j}|}}),\\
			{d_m}^{12} = \arccos (\frac{{o{s_1} \cdot o{s_2}}}{{|o{s_1}| \cdot |o{s_2}|}}).
		\end{align}
	\end{subequations}
	
	If the observed stars $1,2$ and the catalog stars $i,j$ are matched, it is satisfied that:
	
	\begin{equation}\label{d<=ybxl}
		|{d_r}^{ij} - {d_m}^{12}| \leqslant \varepsilon ,
	\end{equation}where $\varepsilon $ denotes the uncertainty of the angular distance calculation.

	\subsubsection{Catalog Star Pairs Database}
	Angular distance feature matching is essentially a process of querying a table, and the first step is to construct a database. According to the sensor performance, the catalog stars with large magnitude are removed. There are $N_{c}$ catalog stars after filtering, and they are combined to obtain ${N_c}({N_c} - 1)/2$ pairs, only those within the field of view are kept.
	The angular distance of each star pair is calculated by (\ref{dr,dm}), and the catalog star pairs database is constructed to store the angular distances and their corresponding number in ascending order.
	
	A hash map is employed to improve the speed of the angular distance search. The hash design space is divided into discrete grids, each corresponding to an index value. Objects are mapped to single or multiple grids based on spatial properties. When querying an object, it only needs to find the grid where it is located instead of traversing the whole design space, thus improving efficiency.
	This work divides all the angular distances in the catalog star pairs database at $n_{x}$ intervals to create a hash table for indexing.
	Calculate the hash value $d_{r}$ for each angular distance and store it with the corresponding number in an interval. The hash function is built as follows:
	
	\begin{equation}\label{fhash}
		{f_{Hash}}({d_r},{n_x}) = INT(\frac{{{d_r}}}{{{n_x}}}),
	\end{equation}where $INT( \cdot )$ represents the function of downward rounding.
	
\begin{figure}[tb]
		\centering
		\includegraphics[width=\columnwidth]{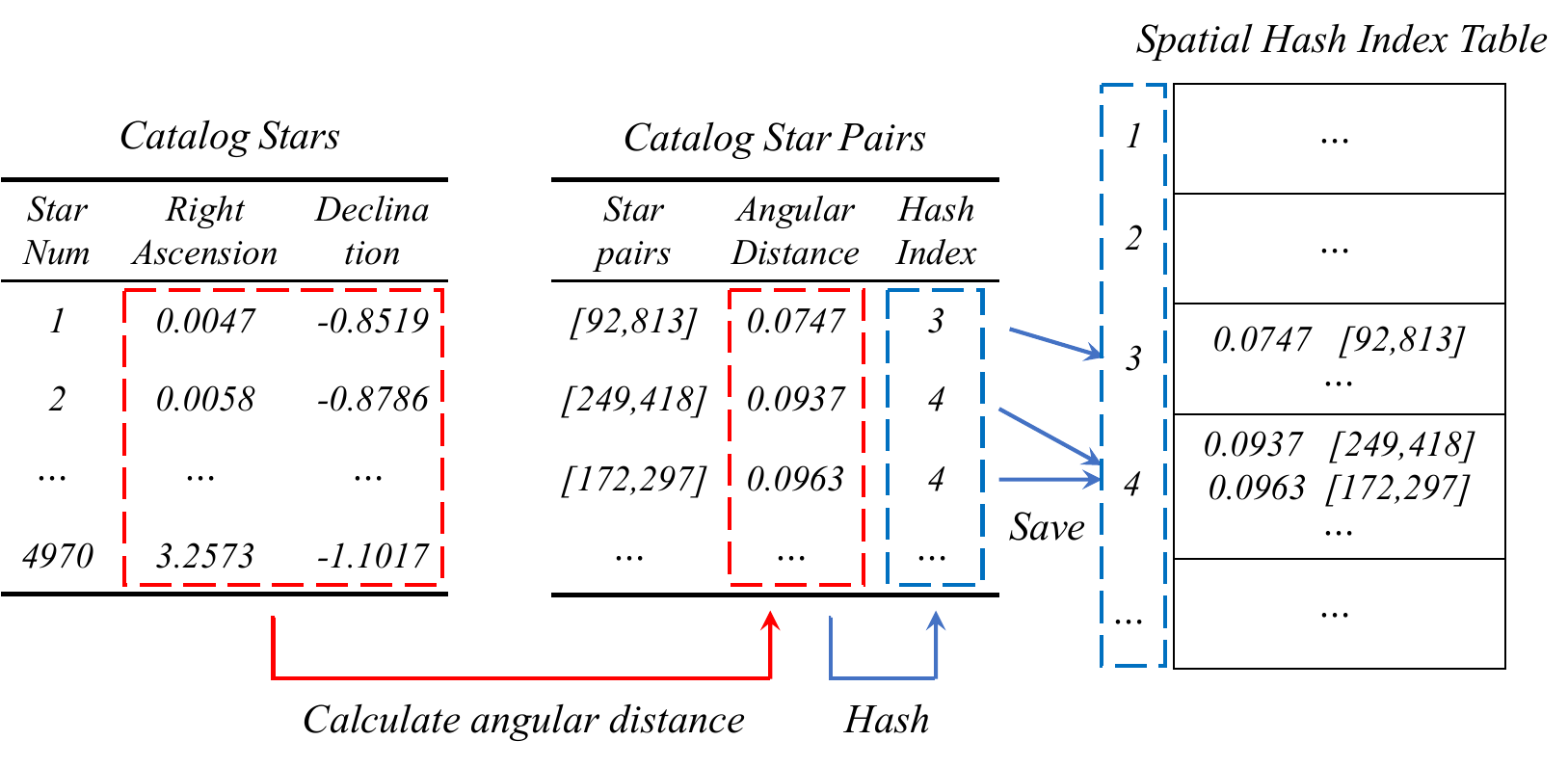}
		\caption{The process of constructing a catalog star pairs database based on spatial hash.}
		\label{fig2}
	\end{figure}

	The tracker will calculate the angular distance by (\ref{dr,dm}) for all star pairs within the observed star map, then calculate the hash interval index value by (\ref{fhash}), thus finding the possible matched catalog star pairs within the corresponding intervals.
	The process of generating a catalog star pairs database based on spatial hash is shown in Fig. (\ref{fig2}).
	
	\subsection{Reverse Attitude Statistics Based Star Map Identification Method}
	\subsubsection{Theoretical Deduction}
	Due to the low feature dimension, matching algorithms based on angular distance inevitably produce redundancy.
	In this case, in an observation frame, the attitude determined by the correct matching is consistent, while the attitude obtained by the wrong matching obeys a random distribution.
	There are $N_{th}$ pairs of stars in the observed map involved in the matching, which corresponds to $N_{th}$ events.
	Event $i$ represents determining the attitude by utilizing the $i_{th}$ pair of observation stars with the initial matched catalog stars.
	Assuming no noise, there must be a correct existence in the initial matching result, \textit{i.e.}, the probability of determining the correct attitude in event $i$ is $P_{r}=1$, and the probability of obtaining a certain wrong attitude is ${P_f} \ll 1$.
	Counting the frequency of attitudes obtained in all events, the frequency of the correct attitude is $F_{r} = N$, and the frequency of a certain wrong attitude is ${F_f} = {P_f} \times N \ll N$, which is derived that:
	
	\begin{equation}\label{Fr>>Ff}
		{F_r} \gg {F_f}.
	\end{equation}
	
\begin{figure}[tb]
		\centering
		\includegraphics[width=1\columnwidth]{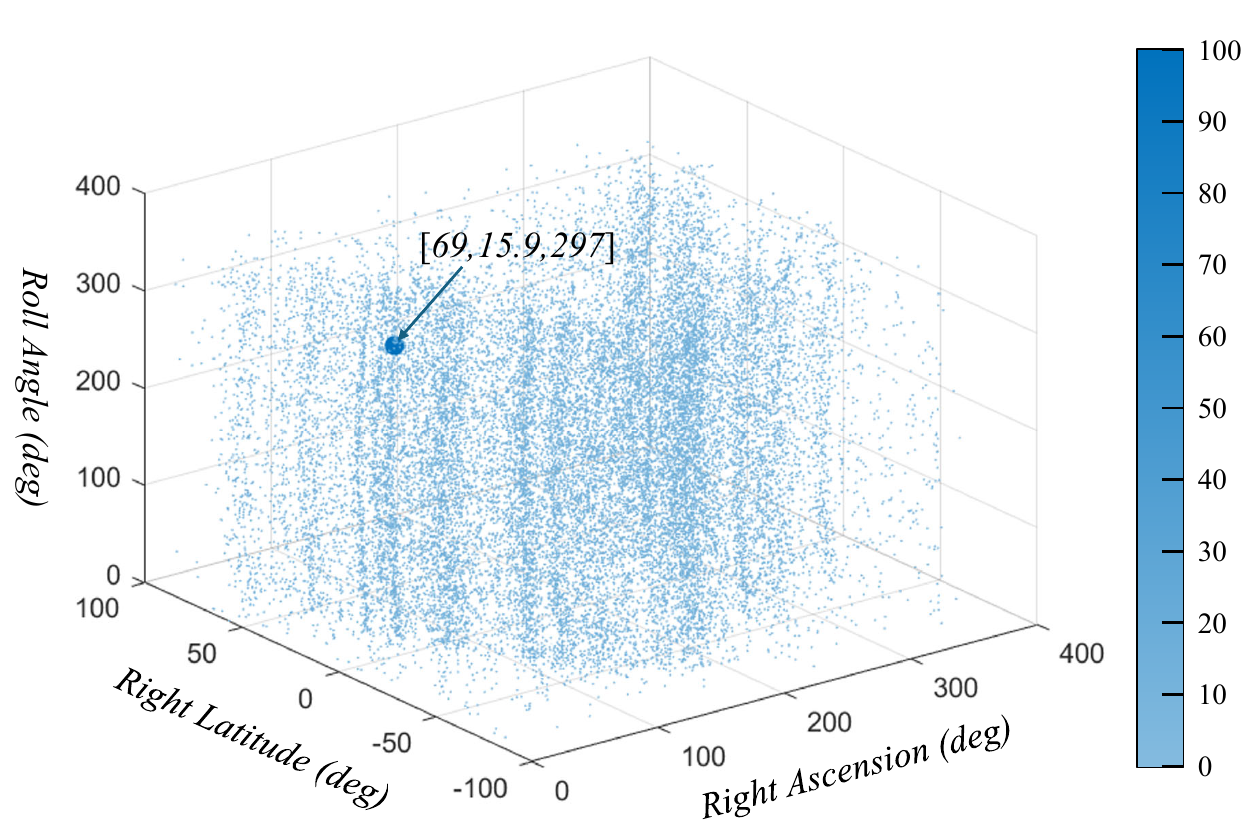}
		\caption{The distribution of potential attitude, where each dot represents a potential attitude, with the size and color depth of the dot being proportional to the frequency.}
		\label{fig3+1}
	\end{figure}

	As shown in Fig. (\ref{fig3+1}), each of these points represents a potential attitude. Therefore, the attitude with the highest frequency is considered to correspond to a correct match.
	
	\subsubsection{Angular Distance Based Initial Matching}
	
	\begin{figure}[tb]
		\centering
		\includegraphics[width=0.75\columnwidth]{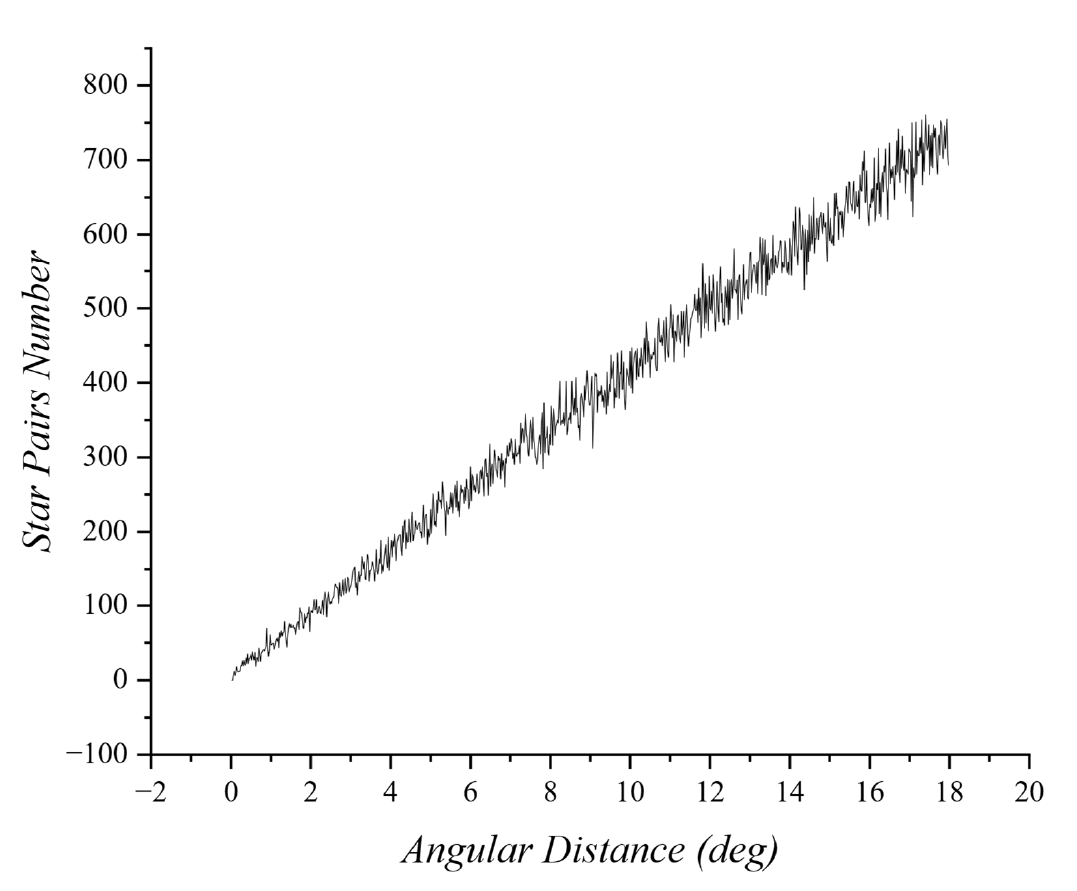}
		\caption{The angular distance distribution of the whole celestial sphere.}
		\label{fig4}
	\end{figure}
	\begin{figure}[tb]
		\centering
		\includegraphics[width=0.85\columnwidth]{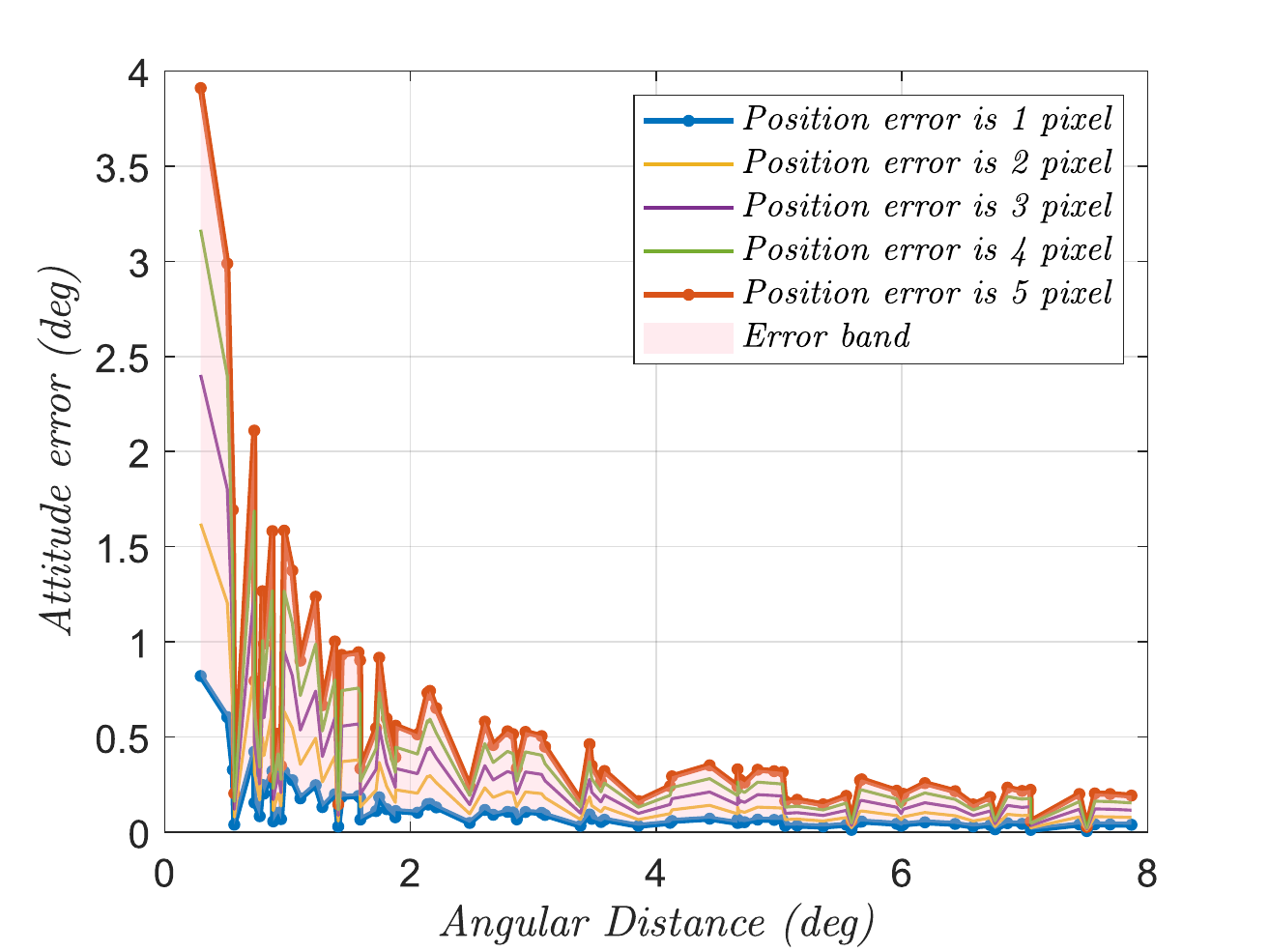}
		\caption{Attitude error caused by star position noise at different angular distances.}
		\label{fig4+1}
	\end{figure}
	
	\begin{figure*}[tb]
		\centering
		\includegraphics[width=0.8\linewidth]{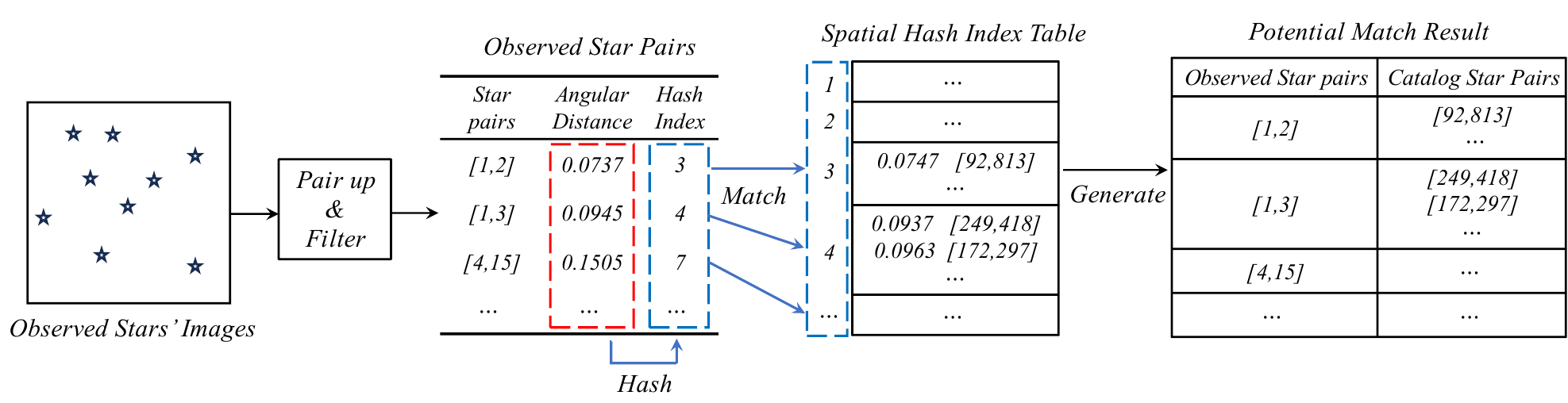}
		\caption{Angular distance features based initial matching algorithm.}
		\label{fig5}
	\end{figure*}

	\begin{figure}[tb]
		\centering
		\includegraphics[width=0.95\columnwidth]{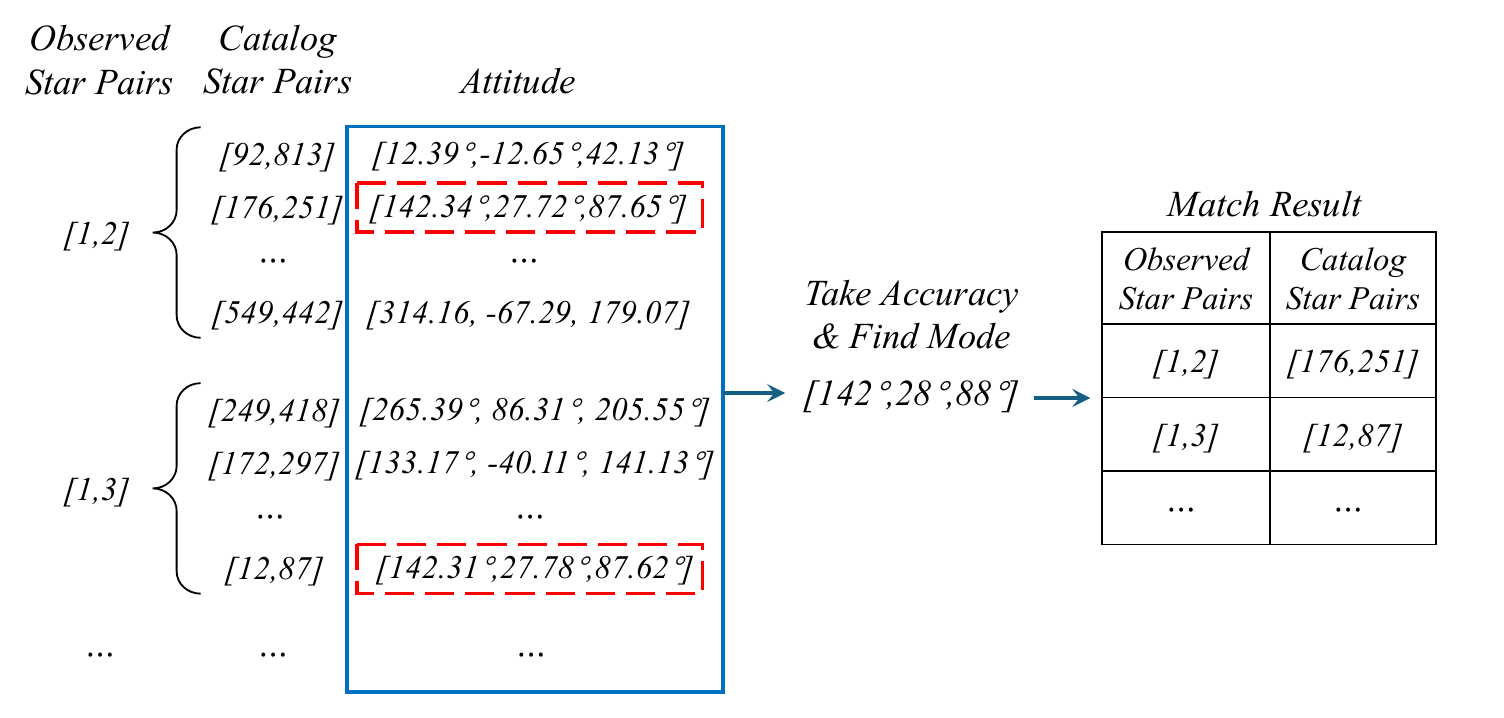}
		\caption{Attitude statistics based accurate matching.}
		\label{fig6}
	\end{figure}
	
	To improve the efficiency and effectiveness of the algorithm, pairs of stars for matching are selected from the observed star map by angular distance.
	There are $N_{os}$ stars in the observed star map, paired as ${N_{op}} = {N_{os}}({N_{os}} - 1)/2$ pairs.
	The maximum condition is set to ensure matching speed under various observation environments. When the number of star pairs exceeds ${N_{th}}$, the algorithm selects ${N_{th}}$ star pairs for matching.
	Principles for selecting star pairs include minimizing redundancy and avoiding distortion:
	
	\paragraph{Minimizing redundancy principle} 
	Fig.(\ref{fig4}) shows the distribution of the angular distance of the entire celestial sphere, and Fig.(\ref{fig4+1}) shows the attitude error caused by star position noise at different angular distances, which indicates that selecting a large angular distance will result in a huge number of redundant matches. However, a small angular distance makes the attitude calculation sensitive to the star position noise, hence it is important to choose an appropriate angular distance.
	When the angular distance is larger than 1.5\textdegree, the error of attitude calculation can be kept within 1\textdegree even if the position noise reaches 5 pixels, which is acceptable for this work. Therefore, when observing stars are sufficient, star pairs with small angular distance values and larger than 1.5\textdegree are preferred for matching.
	
	\paragraph{Avoiding distortion principle}
	Since the distortion is severer in the edge part compared with the image center, the star pairs located in the center region are preferred for matching.
	
	Next, the corresponding spatial hash index values of the observed star pairs are calculated as follows:
	
	\begin{equation}
		{h_{id}}({d_m},{n_x}) = INT(\frac{{{d_m}}}{{{n_x}}}),
	\end{equation}which constructs a hash function and the generated hash index table is a hash map. This results in a complexity of $O(1)$ \cite{spratling2009survey} for the lookup table operation, which significantly improves the efficiency of the initial matching. In this way, pairs of possible matches in the spatial hash index table are found and stored. The process of the initial matching algorithm based on angular distance is shown in Fig.(\ref{fig5}).
	\subsubsection{Dual-vector Attitude Determination Algorithm}
	In this work, reverse attitude statistics is utilized for the accurate match, which needs to calculate the attitudes by the potential match result first.
	The dual-vector-based algorithms are widely applied to determine the initial attitude of star trackers \cite{cilden2015covariance}.
	In the coordinate systems $S_{r}$ and $S_{m}$, the projections of two noncollinear vectors $V_{1}$ and $V_{2}$ are denoted as $V_{1r},V_{1m}$, and $V_{2r},V_{2m}$, respectively.
	The approach to solving the orientation between $S_{r}$ and  $S_{m}$ by projections is known as the dual-vector attitude determination.
	
\begin{figure}[tb]
		\centering{\includegraphics[width=\columnwidth]{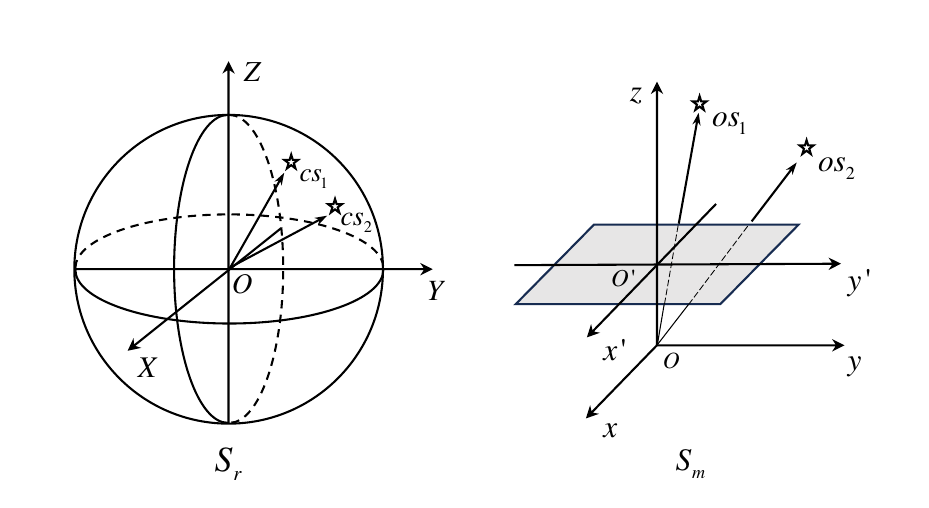}}
		\caption{Dual-vector attitude determination algorithm.}
		\label{fig3}
	\end{figure}

	As shown in Fig.(\ref{fig3}), the vectors of stars 1 and 2 are represented as $cs_{1}, cs_{2}$ in the celestial spherical inertial coordinate system $S_{r}$ ,and $os_{1}, os_{2}$ in the tracker imaging coordinate system $S_{m}$.
	The orthogonal base of $S_{c}$ under $S_{m}$ is given as:
	
	\begin{equation}
		\left\{ \begin{gathered}\label{mabc}
			{m_a} = \frac{{o{s_1}}}{{\left\| {o{s_2}} \right\|}} \hfill \\
			{m_b} = \frac{{o{s_1} \times o{s_2}}}{{\left\| {o{s_1} \times o{s_2}} \right\|}} \hfill \\
			{m_c} = {m_a} \times {m_b}\hfill \\ 
		\end{gathered},  \right.
	\end{equation} the attitude transformation matrix from $S_{c}$ to $S_{m}$ can be obtained as 
	${C_{cm}} = {[\begin{array}{*{20}{c}}
			{{m_a}^T}&{{m_b}^T}&{{m_c}^T} 
		\end{array}]^T}$.
	
	The orthogonal base of $S_{c}$ in $S_{r}$ is given as:
	
	\begin{equation}
		\left\{ \begin{gathered}
			{r_a} = \frac{{c{s_1}}}{{\left\| {c{s_1}} \right\|}} \hfill \\
			{r_b} = \frac{{c{s_1} \times c{s_2}}}{{\left\| {c{s_1} \times c{s_2}} \right\|}} \hfill \\
			{r_c} = {r_a} \times {r_b} \hfill \\ 
		\end{gathered},  \right.
	\end{equation} the attitude transformation matrix from $S_{c}$ to $S_{r}$ can be obtained as ${C_{cr}} = {[\begin{array}{*{20}{c}}
			{{r_a}^T}&{{r_b}^T}&{{r_c}^T} 
		\end{array}]^T}$. 
	
	Due to the relationship between $S_{c}$, $S_{m}$, and $S_{r}$:
	
	\begin{subequations}\label{sc,sm}
		\begin{align}
			{S_c} & = {C_{cm}}{S_m} = {C_{cr}}{S_r},\\
			{S_m} & = {C_{mr}}{S_r},
		\end{align}
	\end{subequations}the attitude transformation matrix $C_{mr}$ from the imaging coordinate system $S_{m}$ to the celestial sphere inertial coordinate system $S_{r}$ can be calculated:
	\begin{equation}\label{cmr}
		{C_{mr}} = {C_{cm}}^{ - 1}{C_{cr}}.
	\end{equation}
	
	Thereby, the attitude of the star tracker is determined as $att_{ij}$.
	
	\subsubsection{Attitude Statistics Based Accurate Matching}
	In the process of accurate matching, to further improve the robustness of the algorithm to noise, the attitudes are processed by uncertainty $fnx$ to obtain $uatt_{ij}$:
	
	\begin{equation}
		{uatt_{ij}} = ({ra_{ij}}^u,{dec_{ij}}^u,{roll_{ij}}^u) = round(\frac{{at{t_{ij}}}}{{fnx}}) \cdot fnx,
	\end{equation}where $i$ denotes the $i_{th}$ pair of observed stars, and $j$ represents the $j_{th}$ pair of catalog stars in the initial matching result, $round( \cdot )$ represents the function of approximate rounding, and ${ra_{ij}}^u,{dec_{ij}}^u,{roll_{ij}}^u$ represent the processed right ascension, right latitude, and the roll angle of tracker, respectively, in units of angle.
	
	Counting the frequency of all attitudes, the one with the highest frequency is regard as the correct attitude:
	
	\begin{equation}
		uat{t_{ij\max }} = {f_{F\max }}(uat{t_{11}},uat{t_{12}},...),
	\end{equation}where ${f_{F\max }}( \cdot )$ donates the function of finding the highest frequency. The $uat{t_{ij\max }}$ corresponds to the correct match, the process is shown in Fig.(\ref{fig6}). This function needs to sort and iterate through the array when finding the mode, and the time complexity of this process is $O(nlogn)$. And the time complexity of calculating the frequency of each attitude and extracting the index is $O(n)$.
	
	\subsubsection{Bayesian Parameter Optimization}
\begin{figure}[tb]
		\centering
		\includegraphics[width=0.8\columnwidth]{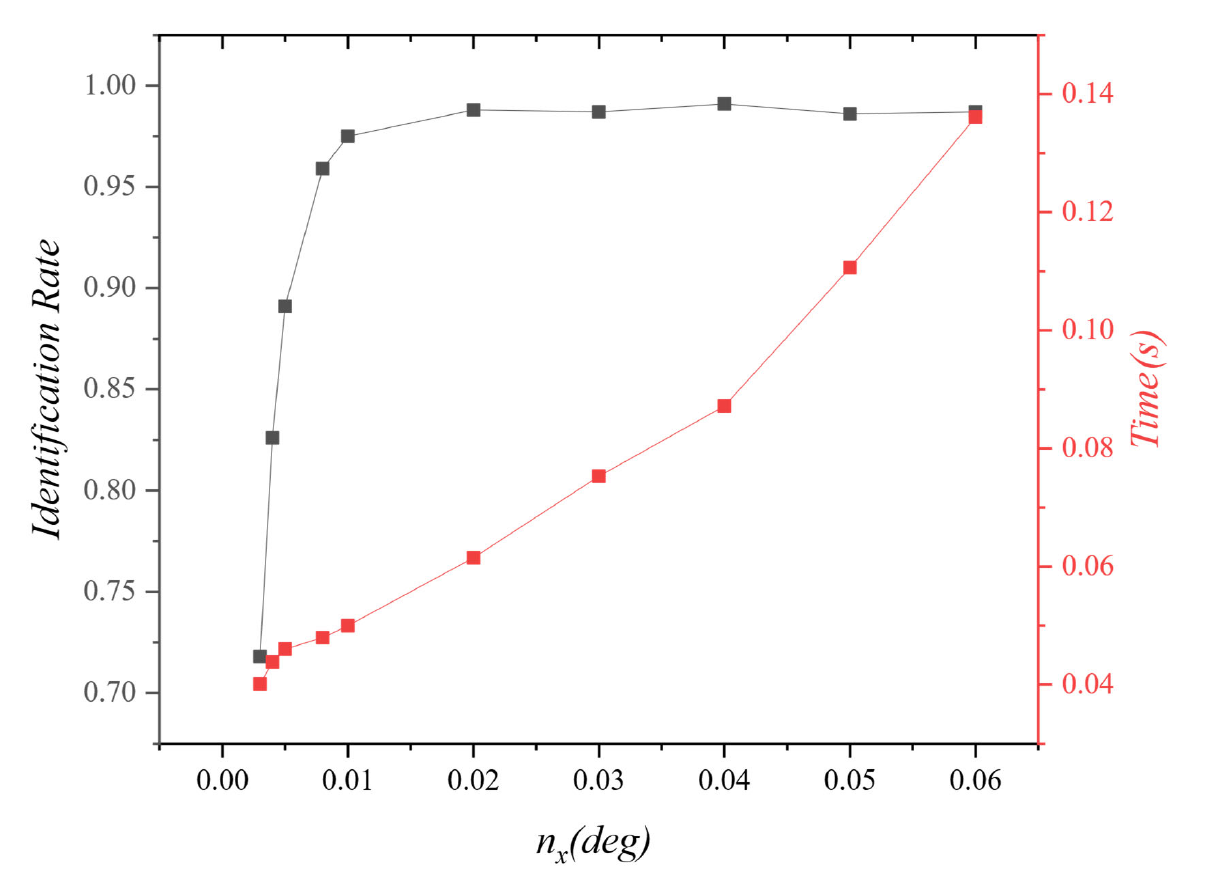}
		\caption{The variations of identification rate and time consumption with $n_{x}$.}
		\label{fig7}
	\end{figure}

	The positions of stars in the observed maps may have errors due to stray light, aerodynamic optical effects, and the self-noise of the tracker.
	This inevitably results in mismatches between the observed and catalog star pairs.
	Therefore, initial matching is applied to improve the robustness of position errors, where the range of angular distances in each interval is determined by $n_{x}$.
	As shown in Fig.(\ref{fig7}), when the position noise is 1 pixel and $N_{th}$ is 20, the identification rate increases as $n_{x}$ rises, which enhances the robustness to noise.
	However, due to the increase in redundant matches, the identification time rises, which affects the efficiency.
	
\begin{figure}[tb]
		\centering
		\includegraphics[width=0.8\columnwidth]{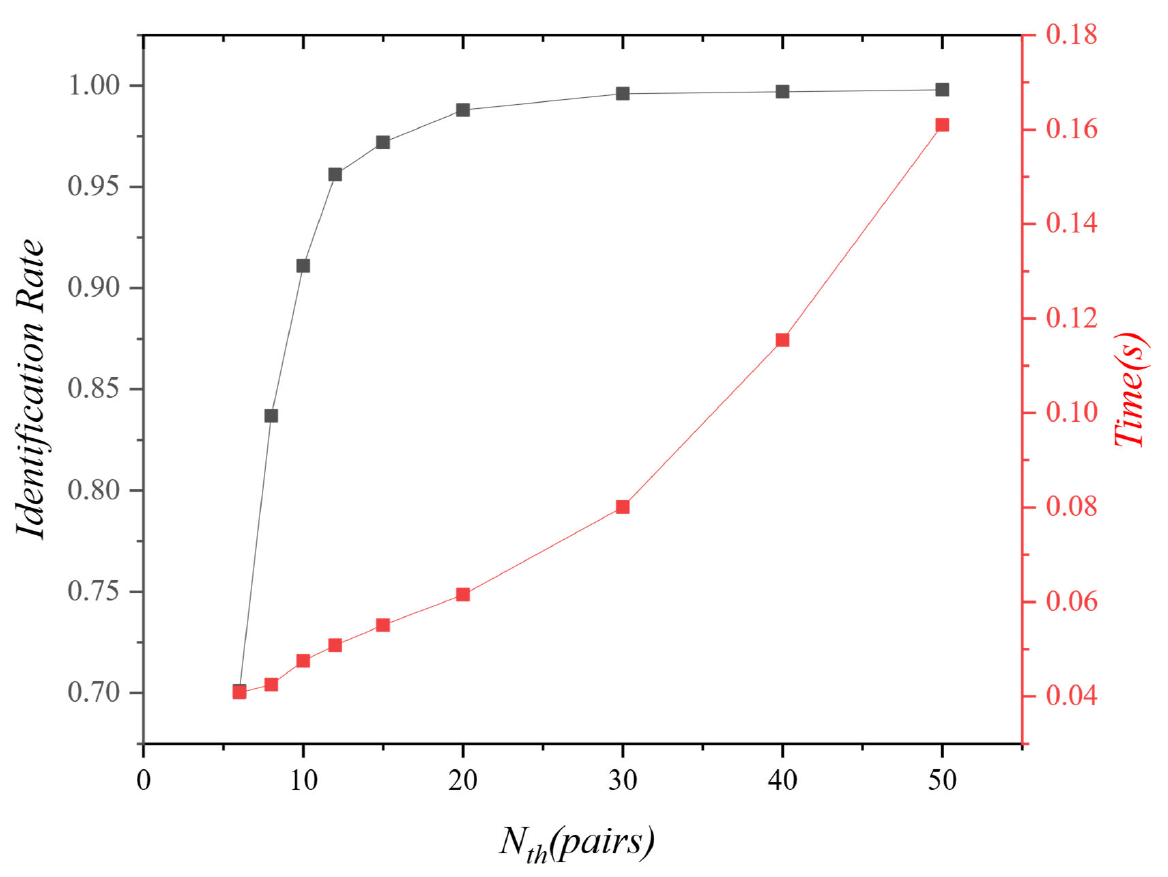}
		\caption{The variations of identification rate and time consumption with $N_{th}$.}
		\label{fig8}
	\end{figure}

	In addition, the identification rate and time spent are also determined by the matching number $N_{th}$.
	When the position noise is 1 pixel and $n_{x}$ is 0.02, the variation of matching time and identification rate with $N_{th}$ is shown in Fig.(\ref{fig8}).
	
	When existing noise, the probability of correctly determining the attitude in each event is less than one.
	If there are few pairs of observed stars in matching, the frequency of the correct attitude may not significantly differ from the others.
	Conversely, the more observed star pairs in a match, the larger the base for statistics. This leads to a higher frequency of correct attitudes and a more reliable result.
	However, excessive star pairs will increase the computational burden and affect the identification speed.
	Therefore, finding proper $n_{x}$ and $N_{th}$ according to the noise contributes to enhancing the performance of algorithms.
	
	Bayesian optimization is a global optimization method which is typically applied to solve the extremum problem for black-box functions \cite{movckus1975bayesian}.
	For the proposed method, the values of $n_{x}$ and $N_{th}$ have a significant impact on the matching time and identification rate, but the relationship between parameters and performance has no explicit expression. The cost function of Bayesian optimization is set as:
	
	\begin{equation}
		\begin{aligned}
			y({n_x},{N_{th}}) = runtime({n_x},{N_{th}}) \times wigh{t_1} \\ 
			- success({n_x},{N_{th}}) \times wigh{t_2} \\ 
		\end{aligned}
	\end{equation}where $runtime$ denotes the running time, $success$ denotes the identification rate, and $weight_{1},weight_{2}$ denote the their weights, respectively.
	Bayesian optimization is employed to find the optimal combination of parameters to minimize the value of the cost function.
	
	\section{Experimental Result}
	
	To evaluate the performance of the proposed method, simulated star maps under different environments are generated. In addition, real star maps from field tests and in-orbit experiments are utilized to validate the robustness further.
	
	\subsection{Parameter Selection}
	A platform is developed to simulate the observed star maps.
	The interaction between the spacecraft and the airflow during its high-speed movement creates a complex flow field, which brings aerodynamic thermal radiation noise to the imaging system, leading to problems such as offset, jitter, and blurring of the stars.
	Additionally, stray light, such as sunlight and atmospheric light, may disturb the tracker, making the signal-noise ratio (SNR) of the stars low and the edges blur, and the imaging system may generate self-noise as well.
	Since the angular distance based star map identification method is mainly affected by star position, star number and false stars, the disturbance of aerodynamic effects and stray light are reflected in the simulation process as position errors, adding false stars and missing stars.
	In this work, the maximum magnitude of catalog stars is 6MV, and the key parameters of the star tracker are shown in table (\ref{Table1}). 
	
	\begin{table}[tb]
		\centering
		\caption{The Key Parameters of Star Tracker.}\label{Table1}
		\begin{tabular}{ll}
			\toprule
			Parameters&Value\\
			\midrule
			Field of view (FOV) & $\Phi8^ \circ$\\
			Focal length & 95mm\\
			Resolution & 2048  $\times$ 2048 pixels\\
			Pixel size & 6.5um\\
			\bottomrule
		\end{tabular}\\
	\end{table}
	
	The parameters of the proposed method are set as follows:
	
	1) $ad_{max}$ represents the maximum angular distance.
	Since the FOV of the simulation platform is 8$^ \circ$, set $ad_{max}=8^ \circ$.
	
\begin{figure}[tb]
		\centering
		\includegraphics[width=0.8\columnwidth]{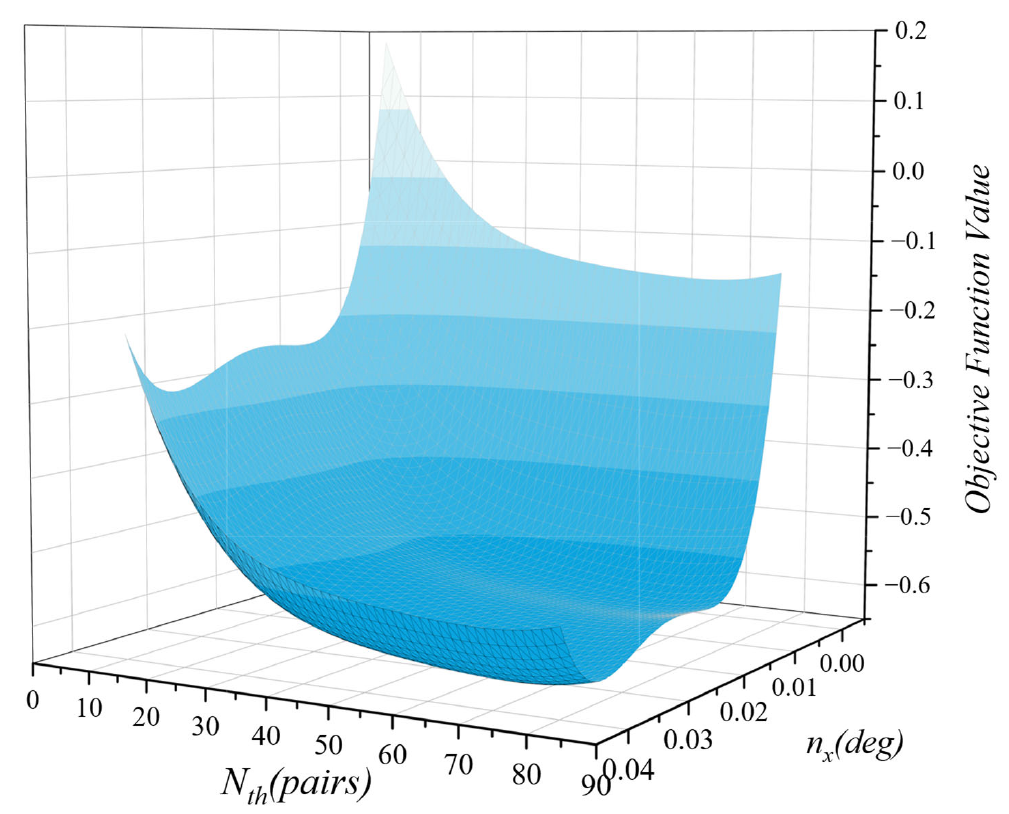}
		\caption{Bayesian optimization result.}
		\label{fig9}
	\end{figure}

	2) $n_{x}$ represents the angular distance range within each interval in the spatial hash table.
	Due to it affecting the robustness of position noise, it is determined by Bayesian optimization.
	As shown in Fig.(\ref{fig9}), when the position noise is 3 pixels, the optimal value of $n_{x}$ is 0.016.
	
	3) The uncertainty of attitude is represented as $fnx$.
	Under the impact of noise, taking a low value may cause correct attitudes to drop out of the original interval. Conversely, taking a high value may result in neighboring false attitudes being counted.
	Therefore, the value of $fnx$ can be adjusted according to the actual situation and is set to 1 in this work.
	
	4) $N_{th}$ represents the upper limit of the observed star pairs involved in the matching.
	More pairs will provide more adequate information, but too many pairs will result in longer running time.
	Therefore, the value is determined by the Bayesian optimization algorithm based on the noise. As shown in Fig.(\ref{fig9}), when the position noise is 3 pixels, the optimal value of $N_{th}$ is 55.
	
	\subsection{Simulation}
	
	The algorithm is validated via a set of simulated star maps with different levels of position noise, false stars, and missing stars. A successful identification of a star map should fulfill the following conditions \cite{wang2017star}:
	
	1) At least two real stars are correctly identified;
	
	2) Neither false stars are identified, nor true stars are incorrectly matched.
	
	To better analyze the performance of the proposed methods, the approaches based on improved triangle \cite{yang2022robust} and hash map \cite{wang2017star} are applied as references.
	Both are based on angular distance and perform well in robustness and identification speed.
	All the methods are executed on a Core I7 PC, the software language used for the algorithm is MATLAB. The results are discussed as follows:
	
	\subsubsection{Position Noise}
\begin{figure}[tb]
		\centering
		\includegraphics[width=0.8\columnwidth]{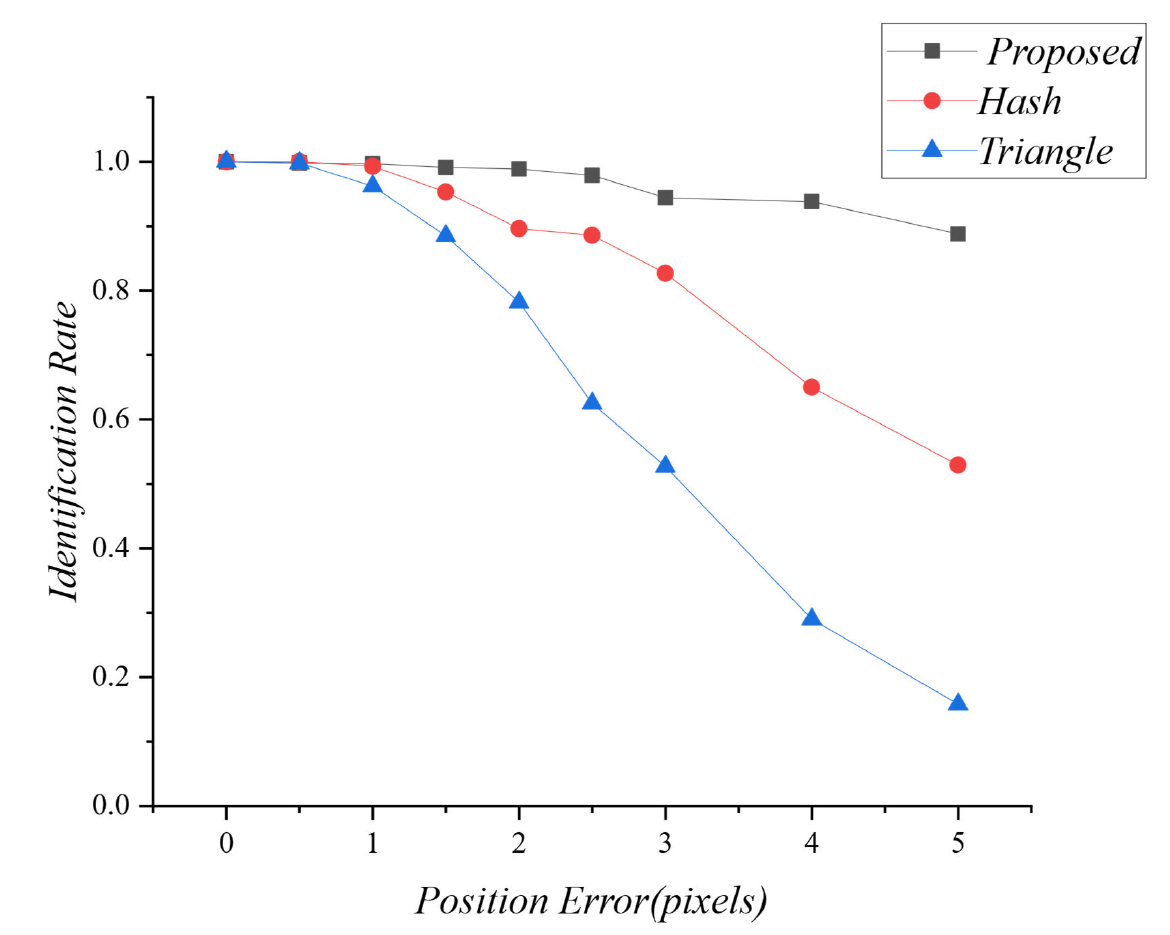}
		\caption{Identification rates for star images in different levels of position noise.}
		\label{fig10}
	\end{figure}

	The position noise in simulation is a random bias obeying a Gaussian distribution with a mean of 0 and a standard deviation of $\sigma$.
	One thousand simulated star maps with $\sigma$ from 0 to 5 pixels are randomly generated for testing, and the average of the identification rate and time are recorded as shown in Fig.(\ref{fig10}) and table (\ref{Table2}).
	When $\sigma$ is 0, the identification rate of all three approaches is 100\%. As $\sigma$ increases, their identification rates all decrease, and the time spent increases.
	Compared with the others, the proposed method has a higher identification rate and spends a shorter time under different levels of $\sigma$.
	It is able to maintain the identification rate of 88.8\% and spend 0.163s even with position noise of 5 pixels, while the approaches based on hash map and improved triangle show underperformance.
	
	\begin{table}[tb]
		\centering
		\caption{Identification Time for Star Images 
			in Different Levels of Positional Noise (Unit: s).
		}\label{Table2}
		\begin{tabular}{lllllll}
			\toprule
			Position error (pixel)&0&1.0&2.0&3.0&4.0&5.0\\
			\midrule
			Triangle&0.172&0.213&0.237&0.352&0.477&0.598\\
			Hash&0.065&0.138&0.199&0.524&0.585&0.617\\
			Proposed&0.046&0.081&0.098&0.093&0.166&0.163\\
			\bottomrule
		\end{tabular}
	\end{table}
	
	\subsubsection{False Stars}
\begin{figure}[tb]
		\centering
		\includegraphics[width=0.75\columnwidth]{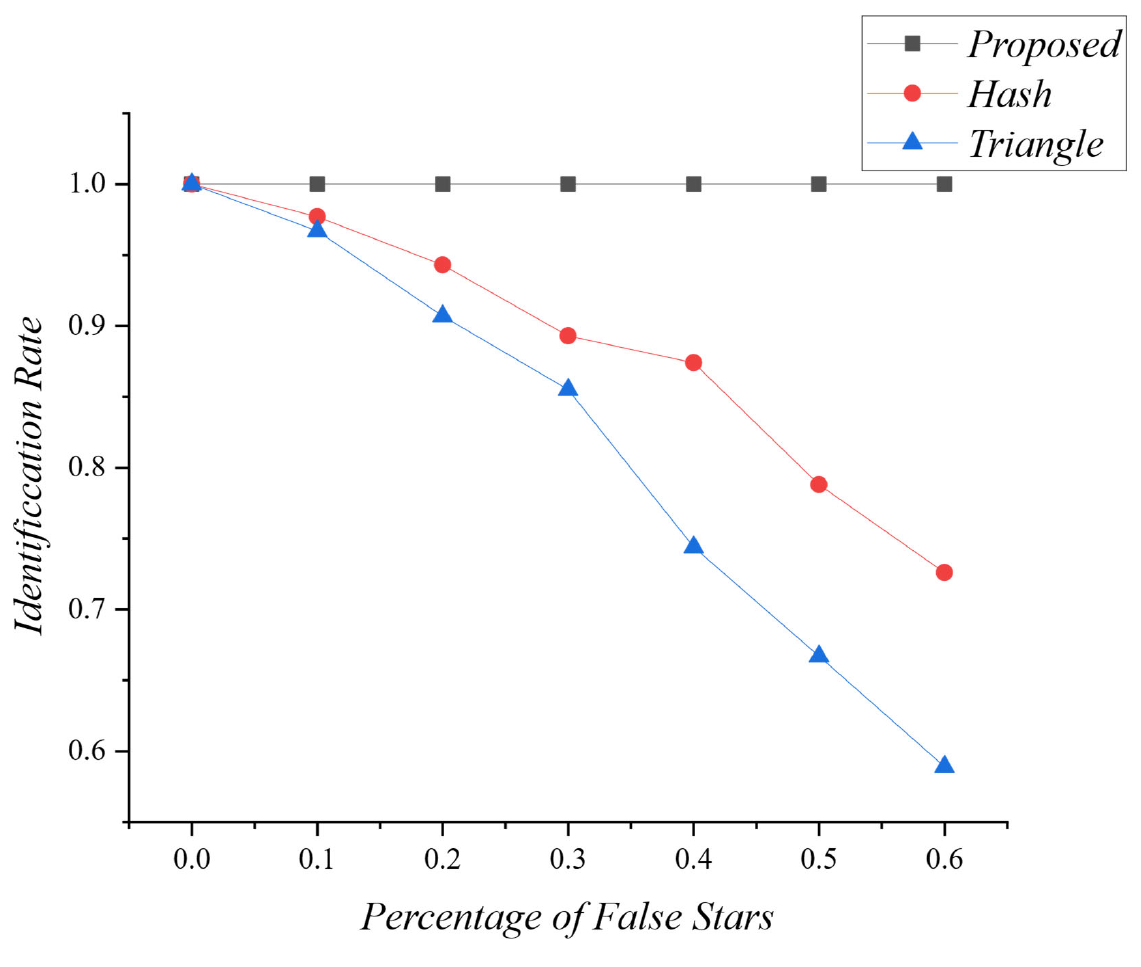}
		\caption{Identification rates for star images with different percentages of false stars.}
		\label{fig11}
	\end{figure}

	The disturbance of false stars in this work is the random addition of stochastic magnitude stars in the observed star map.
	The percentage of added false stars is denoted as $R_{f}$. One thousand simulated star maps with $R_{f}$ of 0 to 60\% are generated, which are utilized to test the identification rate and time spent of the three approaches under different levels of false stars.
	The average results of 1000 star maps are shown in Fig.(\ref{fig11}) and table (\ref{Table3}). When $R_{f}$ is 0, the identification rate of the three algorithms is 100\%. With the increase of $R_{f}$, the identification rates of hash-map-based and improved-triangle-based algorithms are reduced. While the proposed method remains an identification rate of 100\% even under the condition of $R_{f}$ rising to 60\%.
	In terms of time consumption, the proposed method is insensitive to false stars and stays under 0.057s. The time spent by the other algorithms all increases with the false star ratio, reaching 0.795s and 0.456s in the end, respectively.
	
	\begin{table}[tb]
		\centering
		\caption{Identification Time for Star Images 
			With Different Percentage of False Stars (Unit: s).
		}\label{Table3}
		\begin{tabular}{llllllll}
			\toprule
			$False~star$&0&10\%&20\%&30\%&40\%&50\%&60\%\\
			\midrule
			Triangle&0.172&0.204&0.249&0.306&0.328&0.498&0.795\\
			Hash&0.070&0.151&0.133&0.160&0.166&0.412&0.456\\
			Proposed&0.046&0.048&0.048&0.047&0.054&0.057&0.057\\
			\bottomrule
		\end{tabular}
	\end{table}
	
	\subsubsection{Missing Stars}
\begin{figure}[tb]
		\centering
		\includegraphics[width=0.8\columnwidth]{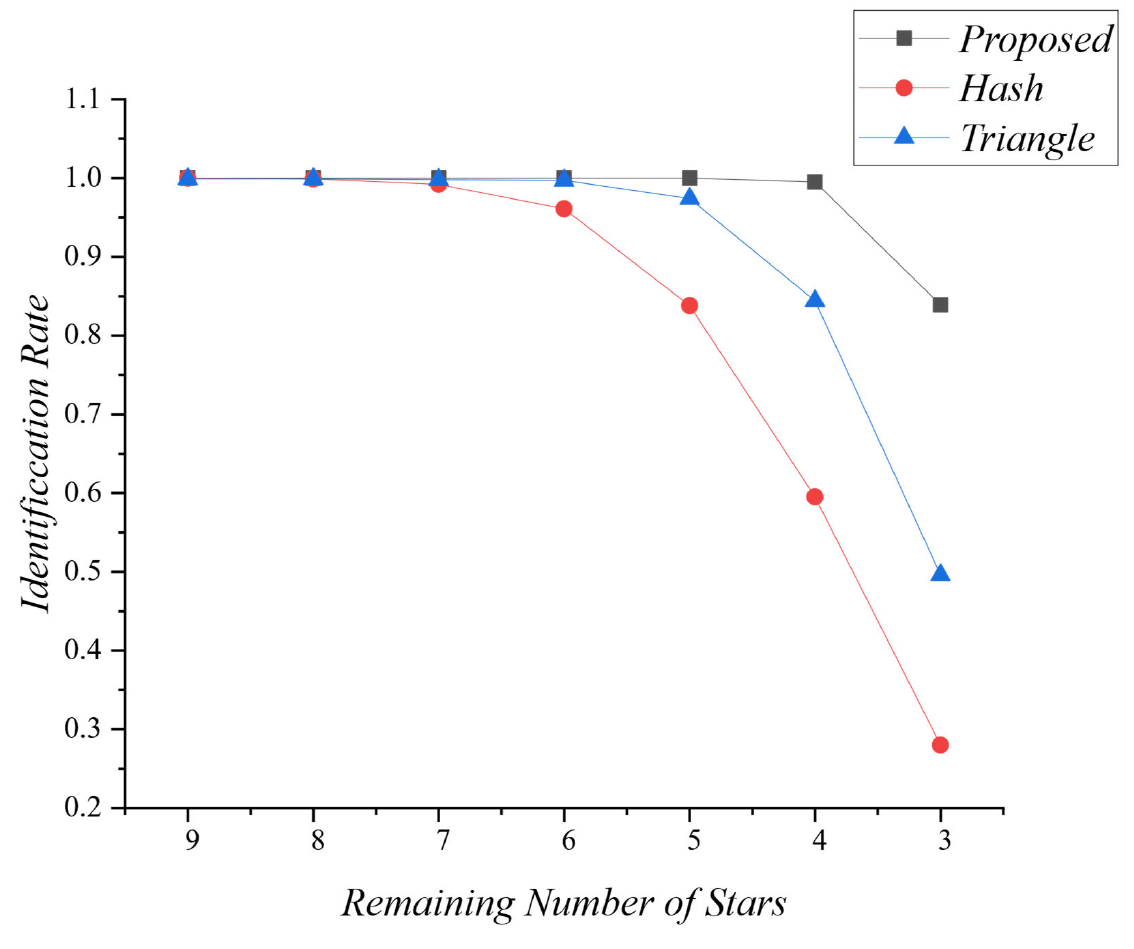}
		\caption{Identification rates for star images with a different number of remaining stars.}
		\label{fig12}
	\end{figure}

	The simulation of the missing star case is a random preservation of stars in the observed star map.
	The number of remaining stars is represented as $N_{r}$. One thousand simulated star maps with $N_{r}$ of 3 to 9 are generated for testing the identification rate and time consumption of the three methods in the case of missing stars, and the average test results of the 1000 star maps are recorded.
	The identification rates of the three algorithms are shown in Fig.(\ref{fig12}), and Table (\ref{Table4}) shows their time spent.
	When $N_{r}$ is 9, the identification rate of three algorithms is 100\%.
	As $N_{r}$ decreases, the identification rate of each method is reduced. Compared with the others, the proposed algorithm shows higher robustness under different $N_{r}$. In particular, its identification rate is 83.9\% with $N_{r}$ of 3, while hash-map-based and improved-triangle-based algorithms have identification rates of 28\% and 49.6\%, respectively.
	The proposed method also exhibits a stable identification time, which is able to stay less than 0.014s with different $N_{r}$. The other time consumption suddenly shortens at $N_{r}$ of 3 due to the fact that some maps can't be matched properly when there are too few stars, which in turn shortens the time.
	
	\begin{table}[tb]
		\centering
		\caption{Identification Time for Star Images 
			With Different Number of Remaining Stars (Unit: s).
		}\label{Table4}
		\begin{tabular}{llllllll}
			\toprule
			$Star~num$&9&8&7&6&5&4&3\\
			\midrule
			Triangle&0.320&0.282&0.262&0.242&0.164&0.498&0.022\\
			Hash&0.087&0.075&0.064&0.057&0.049&0.063&0.008\\
			Proposed&0.015&0.014&0.012&0.010&0.009&0.006&0.003\\
			\bottomrule
		\end{tabular}
	\end{table}
	
	\subsection{Field Test}
\begin{figure}[t]
		\centering
		\includegraphics[width=0.6\columnwidth]{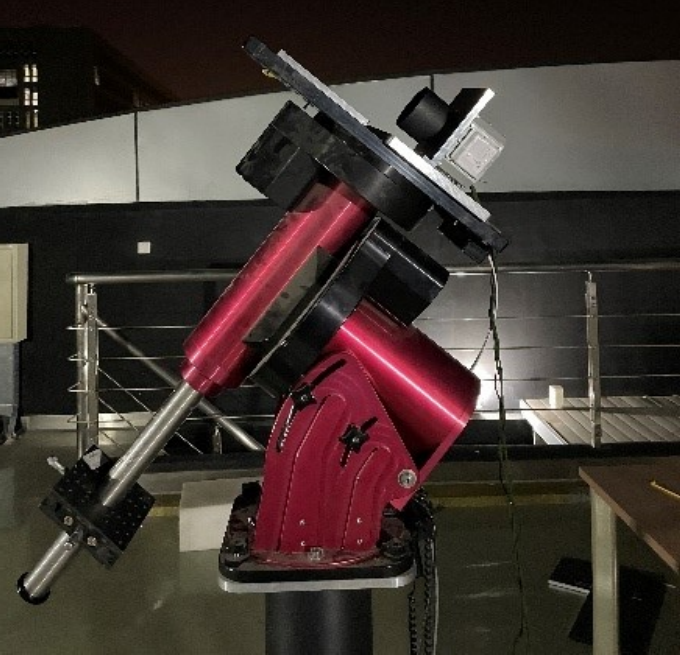}
		\caption{The experimental tracker platform employed in the field test.}
		\label{fig13}
	\end{figure}

	As shown in Fig.(\ref{fig13}), the experimental tracker platform is built to validate the proposed method.
	The key parameters of the tracker are set as table (\ref{Table1}). Over several nights, 148 star maps from different celestial regions are randomly collected. In each region, 10 to 20 static star maps are captured at various angles. Additionally, with the aid of an equatorial mount, 75 dynamic star maps are taken. In total, 1353 star maps are obtained.
	The experiment results indicate that the proposed method successfully identifies all the star maps with an average processing time of 0.059s.
	
	\begin{figure*}[tb]
		\centering
		\includegraphics[width=0.9\linewidth]{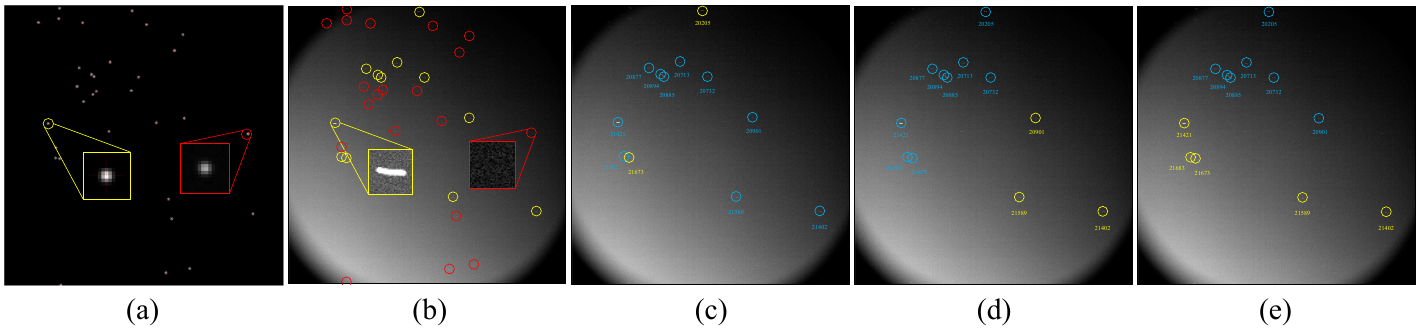}
		\caption{(a)-(b) are the simulation and on-orbit images of the same attitude, respectively, where the yellow circles are stars that can be extracted and the red circles are stars that cannot be extracted due to the influence of environmental light; (c)-(d) are the results of applying the proposed, hash-based, and triangle-based star map identification method, respectively, where the blue circles are successfully matched stars.}
		\label{fig14}
	\end{figure*}
	
	\begin{figure}[t]
		\centering
		\includegraphics[width=0.8\columnwidth]{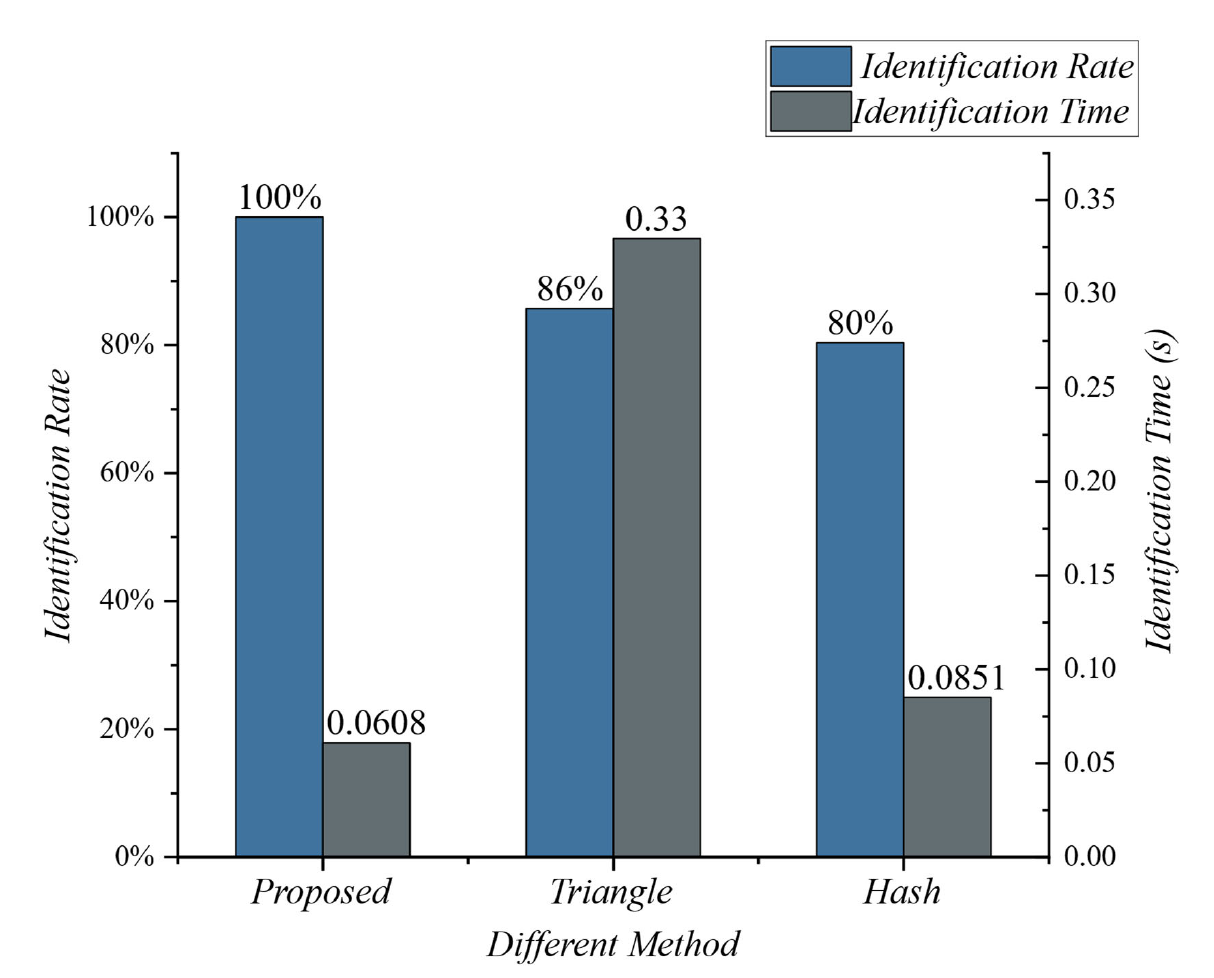}
		\caption{The identification rates and time consumption of different methods.}
		\label{fig15}
	\end{figure}
	
	\subsection{On-orbit Experiment}
	In the on-orbit experiments, a total of 56 on-orbit star maps are collected by the near-space vehicle to validate the proposed algorithm. The key parameters of the tracker are the same as Table (\ref{Table1}).
	The star map of a certain observation and the results of applying different methods of star map identification are shown in Fig.(\ref{fig14}), part of the stars are submerged due to the combined effect of atmospheric background light and aerodynamic environment in near space. Moreover, high-speed maneuvering also leads to the star trailing phenomenon, which further increases the position error of the star.
	The results of applying different methods are shown in Fig.(\ref{fig15}). Compared with the improved triangle algorithm and hash map approach, the identification rate of the proposed method is 14.3\% and 19.6\% higher, respectively. It also has the shortest average identification time per frame.
	
	\subsection{Discussion}
	From the simulation, field test, and on-orbit experiment results, it can be indicated that compared with the hash map algorithm and the improved triangle algorithm, the proposed method shows superior robustness under various levels of position noise, false stars, and missing stars. It also has a higher identification rate and faster speed.
	
	\subsubsection{Robustness Analysis}
	All three algorithms are based on angular distance. When the observed star maps have position noise, the angular distance is directly affected by the deviation of the star position. Therefore, their identification rates decrease as the error increases.
	Since the algorithms based on hash map and triangle mainly rely on angular distance to match, they are significantly affected by the position noise. However, the angular distance is only used for the initial matching in the proposed method. For the incorrectly matched pairs, their attitudes obey a random distribution, which has a weak impact on the attitude statistics.
	
	When there are false stars in the observed star map, since both hash-map-based and improved-triangle-based algorithms are based on triangles. If a false star is included in a set of three stars to form a triangle, they cannot be matched correctly. As the false star ratio increases, the identification rate inevitably decreases.
	In contrast, the proposed algorithm is based on attitude statistics, where false stars only lead to an increase in the number of incorrect attitudes obeying a random distribution. The existence of a certain number of true stars ensures that the frequency of correct attitudes is far higher than others, thus resisting false star interference.
	
	When there are missing stars in the observed star map, the identification rate of the algorithms based on the hash map and triangle is reduced. When there are fewer stars, there are fewer matching triangles, and the identification rate decreases.
	However, for the attitude-statistics-based algorithm, despite the small number of star pairs involved in the matching, the probability of a correct attitude is still 1 for each event, and the others obey a random distribution.
	Therefore, the frequency of the correct attitude is much greater than the others in attitude statistics, thus diminishing the effect of missing stars.
	
	\subsubsection{Real-time Analysis}
	The proposed method is composed of initial matching based on spatial hash and accurate matching based on attitude statistics.
	A spatial hash indexing method is applied in the initial matching, which improves the retrieval speed. 
	Accurate matching based on attitude statistics involves only simple algorithms.
	From the complexity analysis in Section II, it can be seen that the overall time complexity of the proposed algorithm is $O(klogk)$, which is the fastest growing speed in the total process. This is a linear logarithmic time complexity, compared with the square complexity $O(k^2)$ of the traditional algorithms, it is reduced, where $k$ is the number of potential star pairs within the angular distance measurement tolerance.
	The proposed statistics-based method is slightly affected when noise exists. While the others have a sharp increase in the running time as the position noise and false star ratio increase.
	
	\section{Conclusion and Future Work}
	In this work, a reverse attitude statistics based star map identification method is proposed and validated by simulation, field test, and on-orbit experiment.
	The results indicate that the proposed method has strong robustness, a high identification rate, and fast speed.
	Compared with the state-of-the-art, the identification rate is improved by over 14.3\%, and the time consumption is reduced by over 28.5\%. Experiment results demonstrate that the attitude statistics based method can address the position noise, false stars, and miss stars effectively and Bayesian optimization can further enhance the performance. In the future, the proposed method will be applied to identify the near-infrared (NIR) star maps.

\end{document}